\documentclass[conference]{IEEEtran}
\IEEEoverridecommandlockouts

\usepackage{adjustbox}
\usepackage{amsmath,amssymb,amsfonts}
\usepackage[linesnumbered,ruled,vlined]{algorithm2e}
\usepackage{setspace}
\usepackage{algorithmic}
\usepackage{float}
\usepackage{graphicx}
\usepackage{textcomp}
\usepackage[dvipsnames]{xcolor}
\usepackage{multirow}
\usepackage{mdframed}
\usepackage{balance}
\usepackage{booktabs}
\usepackage{tcolorbox}
\usepackage{makecell}
\usepackage{threeparttable}
\usepackage{colortbl}
\usepackage{hhline}
\usepackage{pifont}
\usepackage{listings}
\usepackage{enumitem}
\usepackage{cite}
\usepackage{url}
\usepackage{wrapfig}
\usepackage{tabularx}
\usepackage{array}

\usepackage{tikz}

\newcommand{\parabf}[1]{\noindent\textbf{#1}}

\newcommand{\zq}[1]{\textcolor{black}{\textbf{}#1}}
\newcolumntype{Z}{>{\zq\bgroup}c<{\egroup}}

\definecolor{ggray}{HTML}{eff0f0}
\definecolor{gggray}{HTML}{E8E8E8}
\definecolor{ggggray}{HTML}{BEBEBE}

\newcommand{\eg}{\textit{e.g.,}}

\newcommand{\ourtool}{\textsc{PerfR}}
\newcommand{\crust}{\textsc{c2rust}}
\newcommand{\raw}{\textsc{RawRust}}
\newcommand{\codex}{\textsc{Codex}}
\newcommand{\ablation}{\textsc{PerfR}$_{\mathrm{R}}$}
\newcommand{\refined}{\textsc{Refined}}

\usepackage[utf8]{inputenc}
\usepackage[english]{babel}

\newcounter{finding}
\newcommand{\finding}[1]{\refstepcounter{finding}
 	\vspace{1mm}
	\begin{mdframed}[linecolor=gray,roundcorner=12pt,backgroundcolor=gray!15,linewidth=3pt,innerleftmargin=2pt, leftmargin=0cm,rightmargin=0cm,topline=false,bottomline=false,rightline = false]
		\textbf{Finding \arabic{finding}:} #1
	\end{mdframed}
}

\newcommand{\distance}{8pt}
\begin{document}

\title{Where Does the Performance Go? Understanding and Recovering Performance Regressions in C-to-Rust Translation}

\author{
\IEEEauthorblockN{
Zhiqiang Yuan\IEEEauthorrefmark{1},
Chong Wang\IEEEauthorrefmark{2},
Zhenhao Zhou\IEEEauthorrefmark{1},
Junwei Liu\IEEEauthorrefmark{1}, and
Xin Peng\IEEEauthorrefmark{1}
}
\IEEEauthorblockA{\IEEEauthorrefmark{1}Department of Computer Science, Fudan University, China}
\IEEEauthorblockA{\IEEEauthorrefmark{2}Nanyang Technological University, Singapore}
\IEEEauthorblockA{\{zhiqiangyuan23, zhzhou24, jwliu24\}@m.fudan.edu.cn}
\IEEEauthorblockA{chong.wang@ntu.edu.sg}
\IEEEauthorblockA{pengxin@fudan.edu.cn}
}

\maketitle

\begin{abstract}
Rust is increasingly adopted as a memory-safe successor to C, and transpilers such as \crust{} can migrate entire C projects while preserving their behavior. Preserving behavior, however, does not guarantee preserving performance. We study this gap on 20 real-world C projects and their \crust{} translations. On the eight projects used for analysis, the translated Rust is more than 3\% slower than the original C on 12 of 19 operations, by up to 62.2\%. By comparing execution-cost profiles, optimized LLVM IR, and compiler optimization remarks across each C--Rust pair, we trace these slowdowns to nine recurring performance issues, spanning residual runtime work, missed compiler optimizations, and C-style source patterns, and derive an optimization direction for each. We then present \ourtool{}, an LLM-based approach that encodes these issues as reusable optimization rules to recover the performance of translated Rust. \ourtool{} profiles each project to locate hot functions and matches them against rule-specific patterns. The LLM then checks each rule's rewrite preconditions against the project context before rewriting, and a rewrite is kept only if it passes both functional and performance validation. On 41 operations from twelve held-out projects, \ourtool{} preserves C outputs on all tested workloads and speeds up 40 operations over the \crust{} translations, 18 of which end up faster than the original C. It outperforms the general-purpose coding agent \codex{} on 31 of the 37 operations where both preserve C outputs, and outperforms its own variant without rule guidance on 34 of the 41 operations.
\end{abstract}

\section{Introduction}
C remains widely used in systems and performance-critical software for its low-level control, predictable performance, and minimal runtime overhead~\cite{DBLP:journals/corr/abs-2201-07845, DBLP:journals/corr/PatelR13a}. However, it is prone to memory-safety errors such as buffer overflows and use-after-free, which remain a major source of systems-software vulnerabilities~\cite{fan2020ac,hanley2023rust}. Rust mitigates these risks through ownership and borrowing while retaining low-level performance~\cite{matsakis2014rust,DBLP:journals/tosem/ZhengWZCL24}, which has motivated efforts to migrate legacy C codebases to Rust, including Rust support in the Linux kernel~\cite{Rust4Linux}. As manual migration is costly and error-prone, automated C-to-Rust translation has become an active research area~\cite{c2rust,crown,DBLP:journals/pacmpl/EmreSDH21,DBLP:journals/corr/abs-2405-11514}.

Existing approaches fall into two families. LLM-based translators aim to produce safer, more idiomatic Rust and increasingly use program analysis and project-level context to coordinate translations across components~\cite{DBLP:journals/corr/abs-2503-17741, ptrTrans, DBLP:journals/corr/abs-2412-14234, DBLP:journals/ese/HongR25, DBLP:journals/corr/abs-2409-10506}. However, their outputs are not guaranteed to preserve behavior, a problem that worsens as project scale grows~\cite{ptrTrans,DBLP:journals/corr/abs-2411-13990,DBLP:journals/corr/abs-2503-17741,DBLP:journals/corr/abs-2412-14234}. Rule-based transpilers such as \crust{} instead apply deterministic syntax-directed rules that preserve source behavior, although their output retains C-oriented representations and extensive unsafe code~\cite{c2rust, crown, DBLP:conf/kbse/HongR24,DBLP:journals/pacmpl/EmreSDH21,DBLP:journals/pacmpl/EmreBPSDH23}. Owing to this behavioral guarantee, rule-based output has become a common starting point for practical migration: it has been used in industrial projects~\cite{Curl}, and recent hybrid approaches refine \crust{}-generated Rust with LLM-guided rewriting~\cite{PR2, DBLP:journals/corr/abs-2506-01427, DBLP:journals/corr/abs-2505-10708}.

Recently, the performance of translated code has drawn attention~\cite{xia2023empirical, DBLP:conf/ndss/LiWLSK25, DBLP:journals/corr/abs-2602-00840}. Behavioral preservation does not imply performance preservation, and regressions after migration can offset the benefits of Rust in performance-critical software, where runtime efficiency is often a key reason for using C. The concern is most direct in rule-based output, which retains many C-style code patterns~\cite{DBLP:journals/corr/abs-2506-01427, DBLP:conf/kbse/HongR24} that can introduce runtime overhead. Xia et al.~\cite{xia2023empirical} evaluate \crust{}-generated programs on a synthetic C corpus and identify Relooper-induced code duplication as one source of runtime overhead, and RUSTY~\cite{DBLP:conf/qrs/HanHWZ22} avoids this control-flow expansion through a specialized translation scheme. General-purpose optimizers such as SemOpt~\cite{ZhaoEtAl2026SemOpt} and MoST~\cite{ZhaoEtAl2026MoST} instead guide LLM-based rewriting with reusable optimization knowledge.

These efforts leave translation-induced performance degradation only partially understood: existing measurements rely on synthetic programs and a single mechanism, while general optimization knowledge is not derived from the mechanisms that paired C--Rust analysis reveals. Consequently, three questions remain open: (1) \textit{whether and how much performance degrades after C-to-Rust translation in real-world projects}; (2) \textit{which execution costs and code-level mechanisms explain such degradation}; and (3) \textit{whether recurring mechanisms can be distilled into reusable rules for automated performance recovery}.

To answer these questions, we conduct an empirical study of performance divergence in rule-based C-to-Rust translation using \crust{}~\cite{c2rust}, a transpiler widely adopted in research~\cite{DBLP:journals/pacmpl/EmreSDH21,DBLP:conf/icse/LingYWWCH22} and industry~\cite{Curl}. Its deterministic translation enables controlled paired C--Rust performance comparison. We construct a benchmark of 20 real-world C projects and their \crust{} translations, using eight projects as the Study Dataset for empirical analysis and reserving the remaining twelve for evaluation. We measure performance at the granularity of \emph{operations}, each being a complete program task with checkable output that may span multiple API calls, such as compression or decompression in \textsc{bzip2}. Across 19 operations in the Study Dataset, the translated Rust incurs more than $3.0\%$ runtime overhead over C on 12, eight of which exceed $25.0\%$ with a maximum of $62.2\%$.
Execution-cost decomposition reveals three recurring profiles: additional dynamic instructions, higher cycles per instruction, or both. We trace these differences through optimized IR, compiler optimization remarks, and translated Rust source, identifying nine recurring performance issues. These issues span residual runtime work, missed compiler transformations, and recurring source-level patterns, each with a corresponding optimization direction.

Building on these findings, we present \ourtool{}, a rule-guided framework for recovering performance lost during C-to-Rust migration. \ourtool{} operationalizes each identified performance issue as an optimization guidance that pairs its \emph{Observed Pattern} with the corresponding \emph{Optimization Direction}.
Given a C project, \ourtool{} first obtains its unmodified \crust{} translation (denoted \raw{}), constructs operation-level functional and performance workloads, and performs pre-optimization normalization. \ourtool{} then localizes rule-specific issues in hot functions using optimized IR, compiler remarks, and Rust source. For each pattern group, the LLM evaluates rewrite preconditions against project context and synthesizes a candidate rewrite using the applicable optimization directions. Each candidate is accepted only if it preserves functionality and yields a net runtime improvement across the synthesized workloads.

We evaluate \ourtool{} on twelve held-out projects against \raw{} and \codex{}, a general-purpose coding agent, using C as the runtime reference. \ourtool{} preserves C outputs on all 41 operations and reduces measured runtime relative to \raw{} on 40. After optimization, 18 operations run faster than C, including 10 by more than 3\%. On \textsc{miniz} checksum computation, for example, overhead changes from $+16.7\%$ for \raw{} to $-66.0\%$ for \ourtool{}. Compared with \codex{}, which produces non-equivalent outputs on four operations, \ourtool{} achieves lower measured runtime on 31 of the 37 operations for which both preserve C outputs. Finally, starting from the same refined baseline, rule-guided optimization achieves lower runtime than unguided optimization on 34 of 41 operations. This paper makes the following contributions:
\begin{itemize}[leftmargin=*]
    \item \textbf{Empirical characterization.}
    We construct a 20-project benchmark and characterize performance divergence through runtime analysis, identifying nine recurring performance issues with corresponding optimization directions.

     \item \textbf{Rule-guided performance optimization.}
We develop \ourtool{}, which turns the identified performance issues into
reusable optimization rules and applies them through hotspot-guided detection,
precondition-aware LLM rewriting, and functional and performance validation.

    \item \textbf{Cross-project evaluation.}
    We evaluate \ourtool{} on twelve projects disjoint from rule
    derivation, compare its performance with C, \raw{}, and \codex{},
    and examine rule applicability and the effectiveness of the
    rule-guided mechanism through an unguided ablation.
\end{itemize}

\section{Background}
\label{sec:background}

\parabf{C and Rust Compilation Pipelines.}
\texttt{clang} and \texttt{rustc} both use LLVM for optimization and machine-code generation, but their frontends can produce different LLVM IR from functionally equivalent C and Rust programs~\cite{LLVM:CGO04,rustc-dev-guide}. These differences arise from language-specific semantics and program representations. They affect both the work expressed in the IR and the information available to LLVM, potentially yielding machine code with different execution costs. LLVM provides two complementary artifacts for examining these differences.
\emph{Optimized IR} reveals what remains after optimization, while \emph{optimization remarks} record selected transformations that LLVM performs or misses, with reasons when available, providing evidence about why optimization outcomes differ. Together, they support compiler-level analysis of C-to-Rust performance divergence.

\parabf{Execution Performance Decomposition.}
For a fixed task and input, CPU execution time follows the performance
equation \(T_{\mathrm{CPU}}=I\times\mathit{CPI}\times T_{\mathrm{cycle}}\),
where \(I\) is the number of retired instructions, \(\mathit{CPI}\) is the
average number of cycles per instruction, and \(T_{\mathrm{cycle}}\) is
the duration of one clock cycle~\cite{HennessyPatterson2017}. When
implementations execute on the same processor under a fixed clock
configuration, execution-cost differences can arise from instruction
count, CPI, or both. Instruction count characterizes dynamic work,
whereas CPI characterizes average per-instruction cost.
Top-Down Microarchitecture Analysis (TMA) complements this decomposition
with a hierarchical view of processor utilization. Its top-level categories
are Retiring, Frontend Bound, Bad Speculation, and Backend Bound, with
Backend Bound further divided into Core Bound and Memory
Bound~\cite{Yasin:ISPASS14}. These categories help identify
microarchitectural bottlenecks associated with CPI differences.

\section{Empirical Study Setup}
  \label{sec:study}

We study the runtime impact  of mechanical C-to-Rust translation using \crust{}~\cite{c2rust} and derive optimization rules for translated Rust. We investigate the following research questions:
\textbf{RQ1 (Performance Divergence):}
How does mechanical C-to-Rust translation affect runtime performance?
\textbf{RQ2 (Cost Decomposition):}
What execution costs underlie the observed performance differences?
\textbf{RQ3 (Performance Issue Patterns):}
What recurring patterns characterize performance issues in translated Rust,
and what are the potential optimization directions?

\subsection{Benchmark Construction}
\label{sec:benchmarks}

\parabf{Project Selection.}
We construct the initial candidate pool from prior project-level C-to-Rust translation studies~\cite{DBLP:journals/pacmpl/EmreSDH21, crown, DBLP:journals/pacmpl/HongR24, DBLP:conf/ndss/LiWLSK25, ptrTrans, PR2, DBLP:journals/corr/abs-2504-15144}. After deduplication, a project is eligible if it (i)~contains at least 1\,KLoC of project-owned C source and headers, excluding tests, examples, and dependencies; (ii)~can be translated by \crust{} into a functionally equivalent crate; and (iii)~supports a deterministic workload exercising a measurable hotspot in project-owned code. We exclude projects whose execution is dominated by I/O or external libraries. These criteria yield 16 projects; to broaden external validity, we add four C libraries satisfying the same criteria: \textsc{xxHash}~\cite{xxHash}, \textsc{lz4}~\cite{lz4}, \textsc{miniz}~\cite{miniz}, and \textsc{http-parser}~\cite{parser}. The resulting benchmark contains 20 projects ranging from 1 to 180\,KLoC, partitioned into an eight-project \emph{Study Dataset} for empirical analysis and performance-issue characterization and a disjoint twelve-project \emph{Evaluation Dataset} for approach evaluation.

\begin{table*}[t]
  \centering
  \caption{Projects and workloads in the Study and Evaluation Datasets.}
  \label{tab:datasets-workloads}

  \scriptsize
  \setlength{\tabcolsep}{4pt}
  \renewcommand{\arraystretch}{1.05}

  \begin{tabularx}{\textwidth}{
      @{}
      l
      r
      >{\raggedright\arraybackslash}X
      >{\raggedright\arraybackslash}X
      @{}
  }
  \toprule
  \textbf{Project}
  &
  \textbf{KLoC}
  &
  \textbf{Operations}
  &
  \textbf{Workload Inputs}
  \\
  \midrule

  \textsc{libcsv}
  & 1.0
  & Parsing; Field escaping
  & 67\,MiB CSV\textsuperscript{\ddag}; 10K escaped fields\textsuperscript{\ddag}
  \\

  \textsc{libogg}
  & 2.9
  & Stream encoding; Decoding
  & 300K / 30K packets (1\,KiB each)\textsuperscript{\ddag}
  \\

  \textsc{xxHash}
  & 3.9
  & XXH64; XXH3; XXH32 hashing
  & 64\,KiB buffer\textsuperscript{\ddag}
  \\

  \textsc{libzahl}
  & 5.7
  & Multiplication; Long division; Modular exponentiation
  & 4,096-bit factors\textsuperscript{\ddag}; 8,192/4,096-bit dividend/divisor\textsuperscript{\ddag};
    4,096-bit base/modulus and 256-bit exponent\textsuperscript{\ddag}
  \\

  \textsc{bzip2}
  & 5.9
  & Compression; Decompression
  & Silesia files\textsuperscript{\dag} (${\sim}$100\,MiB); Derived \texttt{.bz2}
    files\textsuperscript{\ddag} (${\sim}$24\,MiB)
  \\

  \textsc{binn}
  & 6.4
  & Serialization; Deserialization
  & 100K-element \texttt{int32} list\textsuperscript{\ddag}
  \\

  \textsc{json-c}
  & 6.9
  & Parsing; Serialization
  & Canada/CITM/Twitter JSON\textsuperscript{\dag} (4.4\,MiB)
  \\

  \textsc{heman}
  & 13.8
  & Distance field; Lighting; Terrain generation
  &  Grayscale grids\textsuperscript{\dag} ($1024^2$/$2048^2$; $256^2$);
  Terrain\textsuperscript{\ddag} ($512^2$)
  \\

  \midrule

  \textsc{libopenaptx}
  & 1.3
  & aptX encoding; aptX decoding; Stream recovery
  & PCM signal (2\,MiB)\textsuperscript{\ddag}; Derived aptX/aptX-HD streams\textsuperscript{\ddag}
  \\

  \textsc{http-parser}
  & 2.5
  & Request, response, incremental, and URL parsing
  & Request, response streams (18.6K)\textsuperscript{\ddag}; 32K URLs\textsuperscript{\ddag}
  \\

  \textsc{fzy}
  & 2.6
  & Fuzzy scoring; Choice-list search; Candidate-list parsing
  & 11K repository paths\textsuperscript{\dag}; Six queries\textsuperscript{\ddag}
  \\

  \textsc{zopfli}
  & 2.9
  & DEFLATE compression
  & Silesia slices\textsuperscript{\dag} (48--96\,KiB)
  \\

  \textsc{lil}
  & 3.6
  & Script execution; Embedding API
  & 24 upstream scripts\textsuperscript{\dag}; Fixed API inputs\textsuperscript{\ddag}
  \\

  \textsc{miniz}
  & 4.2
  & Compression; Decompression; Streaming; Checksums; ZIP processing
  & Silesia files\textsuperscript{\dag} (48\,KiB--9.7\,MiB); Derived zlib/DEFLATE
    streams and ZIP archives\textsuperscript{\ddag}
  \\

  \textsc{lz4}
  & 5.9
  & Compression; Decompression; Streaming; Dictionary modes
  & Silesia files\textsuperscript{\dag} and stress inputs\textsuperscript{\ddag}
    (48\,KiB--9.7\,MiB); Derived LZ4 streams\textsuperscript{\ddag}
  \\

  \textsc{lodepng}
  & 5.9
  & PNG decoding; PNG encoding; Color conversion; zlib processing; Chunk processing
  & PNG images\textsuperscript{\dag} and derived color variants\textsuperscript{\ddag}
    (1.3--1.8\,Mpx); RGBA buffers\textsuperscript{\ddag}
  \\

  \textsc{libqrencode}
  & 6.7
  & QR encoding; Micro-QR encoding; Structured append
  & URL and text inputs\textsuperscript{\dag}; Identifier payloads\textsuperscript{\ddag} (58 total)
  \\

  \textsc{optipng}
  & 9.2
  & PNG optimization; Format conversion
  & PNG images\textsuperscript{\dag}; Converted variants\textsuperscript{\ddag}~ (13\,KiB--3.8\,MiB)
  \\

  \textsc{brotli}
  & 16.0
  & Compression; Decompression; Streaming; Dictionary modes
  & Silesia files\textsuperscript{\dag} (96\,KiB--9.7\,MiB);
    1\,MiB incompressible data\textsuperscript{\ddag}; Derived Brotli streams\textsuperscript{\ddag}
  \\

  \textsc{libxml2}
  & 180.0
  & Parsing; Validation; XPath Evaluation; Serialization; Dictionary/URI/string utilities
  & Upstream XML and HTML documents (105) and schemas (154)\textsuperscript{\dag};
    Fixed utility inputs\textsuperscript{\ddag}
  \\

  \bottomrule
  \end{tabularx}

  \vspace{1pt}
  \begin{minipage}{\textwidth}
  The upper/lower blocks comprise the Study/Evaluation datasets (8/12 projects); $\dagger$: existing corpora tests;
$\ddagger$: study-generated inputs.
  \end{minipage}

  \end{table*}

\parabf{Operation Selection.}
We define an \emph{operation} as a semantically complete end-to-end task that exercises a coherent unit of functionality, may span multiple APIs, and yields a checkable output. Operations provide a suitable granularity for performance comparison: project-level measurements are too coarse to distinguish functional paths, whereas API-level measurements are too fine-grained to represent complete functional tasks. We select an operation if it (i)~represents core functionality, (ii)~exercises a distinct performance-relevant implementation path, and (iii)~can be executed on both C and the \crust{} translation with outputs checked for equivalence. Multiple operations from the same project are retained only when they exercise distinct implementations. For example, we include both compression and decompression in \textsc{bzip2} because exercise distinct implementation paths, and XXH64, XXH3, and XXH32 in \textsc{xxHash} because they implement distinct hashing algorithms.

\parabf{Workload Inputs.}
For each benchmark operation, we prefer established corpora or upstream-provided inputs, such as the Silesia corpus~\cite{silesia-corpus} and the Native JSON Benchmark~\cite{nativejson-benchmark}. When suitable upstream inputs are unavailable, we generate inputs deterministically from fixed parameters so that the C implementation and its \crust{} translation process identical data. For input-sensitive operations, we use multiple sizes or formats. For inverse operations, we generate the corresponding forward output once with the C reference and reuse it unchanged for both implementations. Overall, the workloads span upstream, generated, and derived inputs across multiple sizes and formats, while preserving identical inputs for each C--Rust comparison.
The selected workloads achieve 73.0--100.0\% coverage of reachable functions in each project.

\subsection{Experimental Settings}
\label{sec:setup}

\parabf{Build Configuration.}
We use \crust{}~0.22.1 to translate the C project into the corresponding Rust. For all comparisons, we compile C with \texttt{clang-17} (17.0.6) and all Rust with \texttt{rustc}~1.77.0-nightly (2024-01-15), both using LLVM~17.0.6. C uses \texttt{-O3 -flto}; Rust uses \texttt{opt-level=3}, fat LTO, one codegen unit, and \texttt{panic=abort}. Both sides target the native CPU with AVX-512 disabled to ensure ISA parity; binary inspection confirms that neither emits AVX-512 instructions.

\parabf{Execution Environment.}
Measurements run on an Intel Core i9-10980XE CPU with 18 physical cores. To reduce system noise, we pin each run to an isolated physical core. We use the \texttt{performance} governor with turbo boost disabled to stabilize frequency, and Intel Resource Director Technology to reduce memory-bandwidth interference.

\parabf{Measurement Protocol.}
For each operation, corresponding C and Rust drivers invoke the respective implementations on the same workload input, and we measure them in pairs to obtain stable and directly comparable runtime estimates. When a single execution is too short for reliable timing, we use a fixed number of in-process repetitions to reach a minimum measurement duration, with the same repetition count for C and Rust. After warm-up, we collect paired measurements while alternating the C--Rust execution order across pairs to reduce temporal drift and order effects. We assess stability using the coefficient of variation (CV) of the pairwise
Rust-to-C runtime ratios and repeat any unstable batch. For accepted batches, we remove ratio outliers using Tukey's $1.5\times$ IQR rule~\cite{tukey1977exploratory} and report median runtimes. A pilot study on representative workloads showed that runtime estimates stabilized with a minimum measurement duration of 500\,ms, five warm-up rounds, and 30 measurement pairs; we therefore use these settings throughout, consider a batch unstable when CV exceeds 3\%, and allow at most two retries.

\parabf{Evaluation Metrics.}
We measure runtime using end-to-end wall-clock time. Let $t_C$ and $t_R$
denote the median execution times of C and Rust for the same operation and input. Relative runtime overhead is $\mathit{overhead}=(t_R-t_C)/t_C$, where positive and negative values indicate slowdowns and speedups, respectively.
To explain runtime differences, we decompose \(T_{\mathrm{CPU}}\) into dynamic work and per-instruction cost using \(R_{\mathrm{inst}}=I_R/I_C\) and \(R_{\mathrm{CPI}}=\mathit{CPI}_R/\mathit{CPI}_C\), respectively.

\subsection{Experimental Procedure}
\label{sec:procedure}

\parabf{RQ1: Performance Divergence.}
Using the settings in Section~\ref{sec:setup}, we compare C with its \crust{}
translation for each operation--input pair and compute runtime overhead.

\parabf{RQ2: Cost Decomposition.}
Separately from runtime measurement, we measure retired instructions and CPU cycles over three runs, using their medians to compute \(R_{\mathrm{inst}}\) and \(R_{\mathrm{CPI}}\).
For operations with materially higher CPI (\(R_{\mathrm{CPI}}\geq1.05\)), we further apply TMA to identify the microarchitectural sources and report the corresponding \crust{} translation and C differences in cycles per instruction.

\parabf{RQ3: Performance Issue Patterns.}
To characterize recurring code-level performance issues, we focus on operations
for which \crust{} incurs more than 3\% runtime overhead over C. For each
operation, we use \texttt{perf} to rank project-owned functions by sampled-cycle
share and select the hottest functions accounting for at least 80\% of sampled
cycles; subsequent analysis is restricted to these functions. We examine the
selected functions using three complementary evidence sources: whole-program
optimized IR, compiler optimization remarks, and translated Rust source.

\emph{Optimized IR.}
We compare whole-program optimized IR of C and its \crust{} translation.
We identify sites where translated Rust retains cost-bearing operations absent
from C, such as bounds-check failure paths or saturating conversions.
We also consider optimizer-relevant information in C but missing from
Rust when its absence is accompanied by additional memory operations;

\emph{Optimization Remarks.}
We compare source-mapped optimization remarks for C and its \crust{} translation. We retain cases where LLVM applies a transformation to C but reports a blocking condition for the corresponding translated Rust code, and discard cases where the reported difference does not persist in the optimized IR;

\emph{Translated Rust Source.}
We use \emph{The Rust Performance Book}~\cite{rustperfbook} as an inspection
guide rather than a predefined taxonomy of performance issues. We derive
concrete inspection criteria from relevant recommendations and apply them to
hot translated Rust code. For example, guidance on heap allocation motivates
inspection of repeated allocation and non-amortized buffer growth. We also
examine recurring structures such as loop-invariant dispatch, pointer--length
memory operations, raw-pointer traversal, and element-wise processing.

We analyze the evidence in three steps. First, we record each performance-relevant candidate instance with its source region in the translated Rust, its manifestation in IR, compiler remarks, or source code, and its cost mechanism. Evidence from multiple representations referring to the same source construct and mechanism is merged. We then discard candidates whose code is not exercised by the workload (58) or belongs to the benchmark harness rather than the project (14), whose evidence does not support the stated mechanism (48), or whose C--Rust difference disappears in optimized IR (15). Of 404 candidates, 269 are retained as \emph{instances}.
Second, we group instances that share an underlying performance mechanism and admit a common remediation strategy. We retain only groups supported by at least three instances; each retained group defines a \emph{Performance Issue}. This yields nine issues covering 268 of the 269 instances. The remaining instance concerns a state variable introduced by \crust{} to emulate \texttt{goto}; observed in only one function, it falls below the support threshold.
Third, for each retained issue, we abstract its recurring manifestations into an \emph{Observed Pattern} and derive an \emph{Optimization Direction} targeting the underlying mechanism. For issues identified through IR or compiler remarks, we use paired C--Rust evidence to determine how to eliminate extra work absent from C or restore optimizations realized in C. For source-level issues, we map the underlying mechanism to a recommendation in \emph{The Rust Performance Book}~\cite{rustperfbook}, such as reusing a buffer across calls rather than allocating and freeing it on every call.

\begin{table*}[!t]
  \centering
  \caption{Runtime overhead ($\Delta$) after \crust{} translation; positive values indicate slowdown over C.}
  \label{tab:rq1-overhead}
  \footnotesize
  \setlength{\tabcolsep}{3pt}

  \begin{tabularx}{\textwidth}{
    @{}
    l >{\raggedright\arraybackslash}X
    l >{\raggedright\arraybackslash}X
    l >{\raggedright\arraybackslash}X
    @{}
  }
    \toprule
    \textbf{Project} & \textbf{Operation ($\Delta$)} &
    \textbf{Project} & \textbf{Operation ($\Delta$)} &
    \textbf{Project} & \textbf{Operation ($\Delta$)} \\
    \midrule

    \textsc{libcsv}  & Parsing $+35.1\%$ &
    \textsc{xxHash}  & XXH64 $-0.3\%$ &
    \textsc{binn}    & Serialization $+40.9\%$ \\

                     & Field escaping $+8.3\%$ &
                     & XXH3 $+33.1\%$ &
                     & Deserialization $+27.9\%$ \\

    \textsc{libogg}  & Stream encoding $+2.2\%$ &
                     & XXH32 $+62.2\%$ &
    \textsc{json-c}  & Parsing $+1.9\%$ \\

                     & Decoding $+0.2\%$ &
    \textsc{libzahl} & Multiplication $+37.3\%$ &
                     & Serialization $+2.8\%$ \\

    \textsc{bzip2}   & Compression $+1.4\%$ &
                     & Long division $-0.6\%$ &
    \textsc{heman}   & Distance-field $+27.6\%$ \\

                     & Decompression $+15.5\%$ &
                     & Modular exp. $+10.2\%$ &
                     & Lighting $+29.5\%$ \\

                     & &
                     & &
                     & Terrain generation $+19.0\%$ \\

    \bottomrule
  \end{tabularx}
\end{table*}

\section{Study Results}

\subsection{RQ1: Performance Divergence} \label{sec:rq1}
Table~\ref{tab:rq1-overhead} reports runtime overhead relative to C after \crust{} translation across 19 operations. Overhead ranges from $-0.6\%$ for \textsc{libzahl} long division to $+62.2\%$ for \textsc{xxHash} XXH32. Seven operations remain within $\pm3\%$ of C, whereas twelve are more than $3\%$ slower, including eight with overhead above $25\%$. Performance divergence also varies substantially within the same project. For \textsc{xxHash}, the same 64~KiB input yields overheads of $-0.3\%$ for XXH64, $+33.1\%$ for XXH3, and $+62.2\%$ for XXH32. Similar contrasts appear in \textsc{libzahl}, where long division incurs $-0.6\%$ overhead but multiplication incurs $+37.3\%$, and in \textsc{bzip2}, where compression incurs $+1.4\%$ but decompression incurs $+15.5\%$. These results indicate that the performance impact of C-to-Rust translation is highly operation-dependent, even within the same project.

\finding{Mechanical translation can substantially degrade runtime performance, though the extent varies by operation.}

\subsection{RQ2: Cost Decomposition}
\label{sec:rq2}

\begin{table*}[!t]
  \centering
  \caption{Execution-cost decomposition of the 12 operations with runtime
  overhead above $3\%$. \textit{Cost} denotes increased dynamic work
  (\textit{Work}), increased CPI only (\textit{CPI}), or both (\textit{Both}).}
  \label{tab:cost-decomposition}
  \footnotesize

  \begin{tabular*}{\textwidth}{
    @{\extracolsep{\fill}}
    c l l r r
    @{\hspace{1.5em}}
    c l l r r
    @{}
  }
    \toprule
    \textbf{Cost} &
    \textbf{Project} &
    \textbf{Operation} &
    $\boldsymbol{R_{\mathrm{inst}}}$ &
    $\boldsymbol{R_{\mathrm{CPI}}}$ &
    \textbf{Cost} &
    \textbf{Project} &
    \textbf{Operation} &
    $\boldsymbol{R_{\mathrm{inst}}}$ &
    $\boldsymbol{R_{\mathrm{CPI}}}$ \\
    \midrule

    \textit{Work} &
    \textsc{binn} &
    Serialization &
    $1.530$ & $0.921$ &
    \textit{CPI} &
    \textsc{heman} &
    Lighting &
    $0.935$ & $1.391$ \\

    \textit{Work} &
    &
    Deserialization &
    $1.450$ & $0.899$ &
    \textit{CPI} &
    \textsc{libcsv} &
    Field escaping &
    $0.994$ & $1.089$ \\

    \textit{Work} &
    \textsc{bzip2} &
    Decompression &
    $1.348$ & $0.860$ &
    \textit{Both} &
    \textsc{libcsv} &
    Parsing &
    $1.267$ & $1.067$ \\

    \textit{Work} &
    \textsc{heman} &
    Terrain generation &
    $1.226$ & $0.970$ &
    \textit{Both} &
    \textsc{xxHash} &
    XXH3 &
    $1.270$ & $1.052$ \\

    \textit{Work} &
    \textsc{libzahl} &
    Modular exponentiation &
    $1.123$ & $0.980$ &
    \textit{Both} &
    \textsc{heman} &
    Distance-field pipeline &
    $1.078$ & $1.186$ \\

    \textit{CPI} &
    \textsc{xxHash} &
    XXH32 &
    $0.475$ & $3.417$ &
    \textit{Both} &
    \textsc{libzahl} &
    Multiplication &
    $1.231$ & $1.113$ \\

    \bottomrule
  \end{tabular*}
\end{table*}

The 12 regressions separate into three execution-cost profiles (Table~\ref{tab:cost-decomposition}). Five are \textit{Work} cases, where instruction inflation is partially offset by lower CPI; three are \textit{CPI} cases, where CPI increases without instruction inflation; and four are \textit{Both} cases, where instruction inflation and higher CPI compound. Overall, instruction inflation occurs in nine of 12 regressions, CPI degradation in seven, and both in four. \textsc{bzip2} decompression illustrates the first profile: it executes 34.8\% more instructions but at 14.0\% lower CPI, partially offsetting the additional work. In contrast, the \textsc{heman} distance-field pipeline executes 7.8\% more instructions at 18.6\% higher CPI, compounding the two costs.

For operations with materially higher CPI, TMA further localizes the increase to microarchitectural bottlenecks. In the \textsc{heman} distance-field pipeline, Memory Bound contributes +0.13 cycles per instruction to the CPI increase after \crust{} translation. For \textsc{xxHash} XXH32, Memory Bound and Core Bound contribute +0.50 and +0.48 cycles per instruction, respectively. Thus, CPI degradation does not arise from a single recurring bottleneck; its source varies across operations.

\finding{Translation overhead stems from both additional dynamic work and higher per-instruction cost, with instruction inflation emerging as the more common source.}

\subsection{RQ3: Performance Issue Patterns}
\label{sec:rq3}
\begin{table*}[!t]
\centering
\caption{Recurring performance issues and optimization directions identified in the Study Dataset.}
\label{tab:optimization-rules}
\footnotesize
\setlength{\tabcolsep}{3pt}
\renewcommand{\arraystretch}{0.98}

\begin{tabularx}{\linewidth}{
    @{}
    >{\raggedright\arraybackslash}p{0.13\linewidth}
    >{\raggedright\arraybackslash}p{0.25\linewidth}
    >{\raggedright\arraybackslash}X
    >{\raggedright\arraybackslash}X
    @{}
}
\toprule
\textbf{Performance Issue} &
\textbf{Definition} &
\textbf{Observed Pattern} &
\textbf{Optimization Direction} \\
\midrule

\textbf{Residual Bounds Checks} &
    Bounds checks remain even when established index-range invariants prove them unnecessary. &
Loop invariants establish valid indices, but optimized Rust IR retains checks leading to \texttt{panic\_bounds\_check} that are absent from C IR.
    &

    Express valid iteration bounds through slices or iterators to expose index–length relationships and enable elimination of redundant checks. \\

\textbf{Redundant Conversion Saturation} &
    Saturating float-to-integer conversions remain even when range invariants prove them unnecessary. &

Optimized Rust IR uses \texttt{llvm.fptosi.sat} or \texttt{llvm.fptoui.sat} for range-bounded operands, whereas C IR uses \texttt{fptosi} or \texttt{fptoui}.

&

Avoid unnecessary saturation with \texttt{to\_int\_unchecked} for finite operands whose values, truncated toward zero, fit the target integer type.
    \\

\midrule

\textbf{Missed Loop Vectorization} &
    Loop vectorization achieved in C is lost after translation to Rust. &
    LLVM vectorizes a C loop but leaves its Rust counterpart
    scalar; remarks identify control-flow, call, or
    memory-dependence barriers. &
    Restructure blocking control flow or calls, or expose valid
    dependence information to enable loop vectorization. \\

\textbf{Missed Hot-Callee Inlining} &
    Hot calls inlined in C remain uninlined in translated Rust. &
    A hot call is inlined in C but retained in Rust; IR
    comparison reveals missing inlining attributes, or remarks
    report excessive cost. &
    Restore missing inlining annotations or simplify the callee to
    reduce its estimated inlining cost. \\

\midrule

\textbf{Repeated Invariant Dispatch} &
    Loops dispatch on callbacks or mode values that remain invariant across iterations.
    &
    A loop repeatedly invokes an invariant callback, \eg{}
    \texttt{cb.expect(...)(args)}, or branches on an invariant mode. &
Resolve loop-invariant callback or mode dispatch once before the loop, then execute a specialized loop for the selected target or mode. \\

\textbf{Inefficient Buffer Management} &
Repeated allocation or fixed-increment growth incurs avoidable heap-allocation overhead. &
An internal buffer is repeatedly allocated and freed, or grows
by fixed increments through \texttt{realloc}. &
Reuse buffers to reduce allocations; reserve capacity upfront
or use geometric growth to reduce reallocations.\\

\textbf{Byte-Oriented Memory Operations} &
Byte-oriented processing fails to exploit known type, width, or
length information. &
Typed values use \texttt{memcpy}/\texttt{memset} or hand-written byte-order helpers; known string lengths are recomputed with \texttt{strlen}. &
Express copy and fill through slices, use fixed-width byte-order
conversions, and reuse known string lengths.\\

\textbf{Scalar Element-Wise Processing} &
Element-wise processing leaves parallel or block-wise
computation opportunities unused. &
A loop processes adjacent elements individually (\eg{} \texttt{out[i] = f(in[i])}), or individually updates or compares elements.&
 Process independent elements in groups, or replace per-element updates and comparisons with equivalent block-wise operations. \\

\textbf{Obscured Memory Access Properties} &
Insufficiently exposed access relationships and invariants inhibit
load elimination or hoisting. &
A loop accesses buffers through raw pointers (\eg{} \texttt{ptr.add(i)}), with separately tracked bounds and implicit aliasing relationships. &
Recover valid \texttt{\&[T]}/\texttt{\&mut [T]} views to expose
access properties to LLVM, or hoist loop-invariant loads. \\

\bottomrule
\end{tabularx}
\end{table*}

Table~\ref{tab:optimization-rules} reports the nine recurring performance issues identified by our analysis, organized by their primary manifestation: residual runtime work (two), missed compiler transformations (two), and source-level performance issues (five). These categories are not mutually exclusive, as residual operations or retained calls may also inhibit downstream compiler transformations.

\parabf{Residual Runtime Work.}
Optimized IR reveals two recurring forms of residual work in translated Rust: \emph{Residual Bounds Checks} and \emph{Redundant Conversion Saturation}. \emph{Residual Bounds Checks} persist despite provably valid access ranges. In \texttt{BZ2\_decompress}, a nested traversal writes \texttt{(*s).inUse[(i * 16 + j) as usize]}, where \texttt{i < 16} and \texttt{j < 16} guarantee an index below 256. Yet the Rust IR retains a comparison against 256 and a branch to \texttt{panic\_bounds\_check}, while the corresponding C IR has no such failure path. The translated state-machine control flow prevents LLVM from recovering this range invariant. Expressing the traversal over a bounded slice makes the valid extent explicit and enables
the redundant check to be eliminated. \emph{Redundant Conversion Saturation} persists when operand ranges rule out saturation but the Rust IR retains saturating float-to-integer conversions. In \textsc{heman}, seven such conversions originate from \texttt{fastFloor} after inlining into \texttt{heman\_internal\_generate\_island\_noise}. The Rust IR uses \texttt{llvm.fptosi.sat} or \texttt{llvm.fptoui.sat}, whereas the corresponding C IR uses direct \texttt{fptosi} or \texttt{fptoui}. The operands are finite grid coordinates whose truncated values remain within the target integer range. Once these conditions are established, \texttt{to\_int\_unchecked} eliminates the redundant saturation.

\parabf{Missed Compiler Transformations.}
Paired C--Rust compiler remarks reveal two recurring missed
transformations in translated Rust: \emph{Missed Loop Vectorization} and
\emph{Missed Hot-Callee Inlining}.
The \emph{Missed Loop Vectorization} issue occurs when LLVM vectorizes a C loop
but leaves its translated Rust counterpart scalar. In \textsc{bzip2}, five hot
decode loops exhibit this divergence.
In a representative move-to-front loop that shifts \texttt{mtfa} entries, each array access carries a bounds check that adds a loop exit, and LLVM reports that it could not determine the number of loop iterations. Expressing the shift as a bounded slice operation can remove these checks and enable vectorization.
The \emph{Missed Hot-Callee Inlining} issue occurs when a hot call inlined in C
remains uninlined in translated Rust because of missing inlining attributes or
excessive estimated callee cost. In \textsc{xxHash}, the C implementation forces
\texttt{XXH3\_accumulate\_512\_scalar} to inline, whereas its translated Rust
counterpart lacks the corresponding annotation and retains the call inside the
hot stripe-processing loop. Restoring the missing inlining annotation can
eliminate this barrier.

\parabf{Recurring Source-Level Patterns.}
Source inspection identifies five patterns in translated Rust code: \emph{Repeated Invariant Dispatch}, \emph{Inefficient Buffer Management}, \emph{Byte-Oriented Memory Operations}, \emph{Scalar Element-Wise Processing}, and \emph{Obscured Memory Access Properties}. Unlike residual runtime work and missed compiler transformations, these patterns are characterized primarily by recurring source structures and their associated cost mechanisms. For example, \emph{Inefficient Buffer Management} occurs in \textsc{libzahl}, where \texttt{zmul\_ll} creates and releases four temporary integers at every recursion level, resizing their storage through \texttt{libzahl\_realloc} when needed. Reusing preallocated scratch storage across recursive calls avoids this repeated allocation and resizing overhead.

The \emph{Obscured Memory Access Properties} issue arises when raw-pointer
accesses leave bounds and aliasing relationships implicit, limiting LLVM's
ability to eliminate redundant loads. In \textsc{libzahl}, \texttt{zrsh}
traverses limb arrays through raw-pointer accesses such as
\texttt{*ptr.offset(i)}. The optimized Rust IR contains 73 static load
instructions, compared with 24 in C, and lacks the \texttt{!tbaa} metadata
present in the C IR. Consistently, GVN emits 46 missed-optimization remarks
for Rust, versus 16 for C, frequently citing potentially aliasing writes as
barriers to load elimination. Together, these observations suggest that
obscured access relationships lead to conservative memory-dependence
assumptions that inhibit load reuse. Recovering valid
\texttt{\&[T]}/\texttt{\&mut [T]} views can expose these properties to LLVM
and enable load elimination.

\finding{We identify nine recurring performance issues in translated Rust, spanning residual work, missed compiler transformations, and source-level patterns, with optimization directions.}

\section{Approach}

\begin{figure*}[!t]
    \centering
    \includegraphics[width=0.95\textwidth]{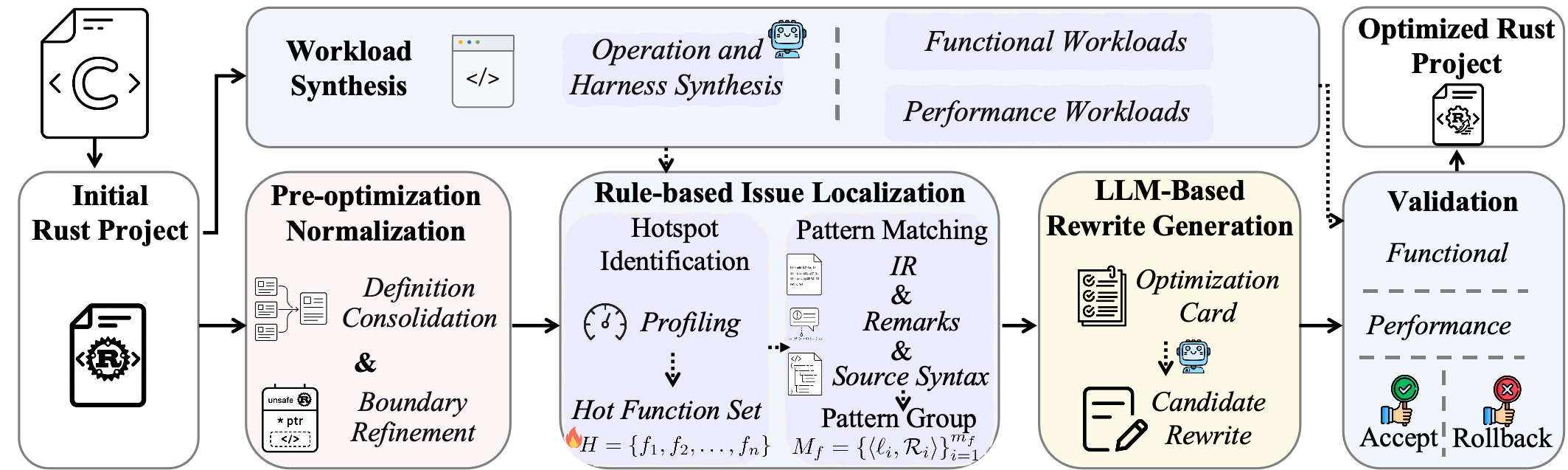}
    \caption{Workflow of \ourtool{}}
    \label{fig:pipeline}
\end{figure*}

The performance issues derived in \S\ref{sec:rq3} provide reusable guidance for recovering translation-induced performance loss. Applying this guidance to a new project requires localizing performance-critical code and identifying applicable rules. We therefore present \ourtool{}, a rule-guided framework for recovering performance lost during C-to-Rust migration. Starting from a C project, \ourtool{} obtains \raw{} with \crust{}, synthesizes workloads, and performs pre-optimization normalization (Figure~\ref{fig:pipeline}). \ourtool{} then localizes hot functions, identifies applicable optimization rules, and uses project context to guide precondition-aware LLM rewrites. Each rewrite is accepted only if it preserves functionality and improves performance.

\subsection{Workload Synthesis}
\label{sec:workload}

Project-level optimization requires executable workloads to locate runtime hotspots and evaluate candidate rewrites. To capture end-to-end behavior and cost, \ourtool{} organizes these workloads around operations that compose related APIs into complete tasks with checkable outputs. Starting from \raw{}, \ourtool{} synthesizes executable operations and harnesses and refines them using coverage feedback. This shared execution basis supports two complementary workload types: functional workloads provide diverse inputs for regression testing, while performance workloads provide sufficiently long executions for hotspot localization and performance validation. These internal workloads are independent of the benchmark workloads used for final evaluation.

\parabf{Operation and Harness Synthesis.}
\ourtool{} first extracts the public API of \raw{}, including \texttt{pub} functions and structs together with their module paths, and provides this information to the LLM. The LLM plans a bounded set of end-to-end operations, each representing a distinct functional task and identifying the API elements needed to realize it. Planning prioritizes core functionality and broad public-function coverage across the operation set.

The planned operations, however, are not directly executable. For each operation, the LLM uses the associated API signatures to generate an executable \emph{harness} and a small set of distinct seed inputs. The harness invokes the relevant APIs for each seed and produces an observable output. \ourtool{} compiles and repeatedly executes the harness to check successful execution and output determinism. It also checks for output variation across distinct seeds to reject degenerate harnesses that ignore their inputs. When a check fails, the corresponding diagnostic feedback is returned to the LLM for revision~\cite{ptrTrans, DBLP:conf/icse/BouzeniaDP25}. This process continues within a fixed repair budget, and an operation is retained only when its harness passes all checks.

Executability alone, however, does not ensure broad coverage of the public API. \ourtool{} therefore measures public-function coverage across the retained operations using an instrumented build and their seed inputs. If coverage remains below the target, the uncovered functions and their signatures are provided to the LLM, which extends existing operations or introduces new ones. The harnesses and seed inputs are then generated and validated through the same process before coverage is recomputed. This feedback loop continues until the coverage target is reached or the refinement budget is exhausted, yielding the final operation set and validated harnesses.

\parabf{Functional Workload Synthesis.}
The validated harnesses establish operation executability, but their small seed sets provide limited behavioral diversity for regression testing. \ourtool{} therefore fuzzes the seeds of each retained operation to generate a broader input corpus. Each generated input is executed repeatedly on the initial \raw{}. Inputs that fail or produce nondeterministic outputs are discarded, while stable output digests are recorded as regression references. The retained inputs, their reference digests, and the corresponding harness together form the \emph{functional workload} used to validate subsequent rewrites.

\parabf{Performance Workload Synthesis.}
Functional workloads prioritize input diversity, but individual executions may be too short for reliable performance measurement. \ourtool{} therefore constructs a separate performance input for each operation by scaling a semantically meaningful input dimension, such as record count, data size, or image dimensions. For operations that consume derived formats, \ourtool{} scales the source data and invokes the corresponding producer operation to construct a valid input. If a single execution remains too short for stable timing, \ourtool{} calibrates a fixed number of in-process repetitions to reach a minimum measurement duration. The harness, performance input, and repetition count together define a \emph{performance workload} used for hotspot localization and performance validation. The operation budget, coverage target, refinement budget, and timing parameters are specified in the experimental setup.

\subsection{Pre-optimization Normalization}
\label{sec:translation-refinement}

Given a C project, \crust{} yields \raw{}, which preserves the behavior of the C program. However, \raw{} retains C-oriented representations, including duplicated definitions, internal \texttt{extern ``C''} boundaries, and pervasive raw-pointer accesses. Although these representations are not the performance issues identified in \S\ref{sec:rq3}, they hinder project-wide analysis and coordinated rewriting while obscuring optimization-relevant program properties. \ourtool{} performs a preparatory refinement that consolidates shared definitions and refines internal ABI, pointer, and safety boundaries. Each refinement is validated against \S\ref{sec:workload} functional workloads; failing changes are reverted.

\subsubsection{Definition Consolidation}

Declarations from shared C headers may appear as duplicate definitions across translated Rust modules. \ourtool{} collects top-level items, including nominal types, type aliases, constants, statics, and functions, and groups them by source provenance. Definitions are compared after ignoring comments and formatting whitespace; attributes, declared types, and item bodies remain part of the comparison. Items originating from the same C declaration with identical normalized definitions form candidate groups. Static definitions remain separate to preserve storage identity. Function consolidation additionally requires consistent resolved dependencies and compatible linkage semantics. Eligible groups are processed in dependency order. For each group, \ourtool{} retains one canonical definition in a shared module and replaces the remaining copies with imports.

\subsubsection{Boundary Refinement}

\parabf{Internal ABI Recovery.}
After definition consolidation, \ourtool{} examines each project-defined
\texttt{extern "C"} function to determine whether the C ABI remains necessary.
The boundary is preserved when the function participates in an external
interface or is used through a C-ABI function pointer. When all uses are
project-local, \ourtool{} removes \texttt{extern "C"} and
\texttt{\#[no\_mangle]}, removes the corresponding external declarations,
and rewrites callers to invoke the function directly through its Rust module
path.

\parabf{Local Pointer-View Recovery.}
After restoring internal Rust call relationships, function bodies may still access objects and buffers through raw-pointer parameters. Following the pointer-semantics recovery strategy of prior C-to-Rust translation work~\cite{ptrTrans}, \ourtool{} analyzes each raw-pointer parameter to determine whether its accesses can be represented by a Rust reference or slice. For a pointer to an individual object, it constructs a local \texttt{\&T} or \texttt{\&mut T} view; for a contiguous buffer, it recovers the accessed extent and constructs \texttt{\&[T]} or \texttt{\&mut [T]}. Shared views require no conflicting writes, while mutable views additionally require exclusive access. When these conditions cannot be established, the raw-pointer representation is retained; otherwise, covered dereferences and pointer-offset accesses are rewritten through the recovered view.

\parabf{Unsafe-Boundary Narrowing.}
Recovering references and slices reduces the code requiring an unsafe context
but may leave caller-side safety obligations. \ourtool{} therefore reassesses
each function-level \texttt{unsafe} annotation. The \texttt{unsafe fn} marker
is retained when callers must uphold safety preconditions that cannot be
established within the function, such as raw-pointer validity. Otherwise,
\ourtool{} removes the marker and confines unsafe operations to local
\texttt{unsafe} blocks; blocks reported as \texttt{unused\_unsafe} by the
compiler are then removed.

\subsection{Rule-based Issue Localization}

Project-level performance optimization requires answering two questions:
where optimization can materially affect end-to-end runtime, and what performance issues should be addressed there. \ourtool{} first profiles the performance workloads to localize project-owned hot functions. \ourtool{} then uses the \emph{Observed Pattern} and corresponding \emph{Optimization Direction} derived for each recurring issue in RQ3~(\S\ref{sec:rq3}) to identify concrete optimization targets within these functions.
\ourtool{} therefore first localizes hot functions and then matches these patterns against their optimized IR, compiler remarks, and Rust source to identify concrete optimization targets.

\subsubsection{Hotspot Identification}
\label{sec:localization}
Project-level optimization need not examine every function equally, since optimizing code with little runtime contribution yields limited end-to-end benefit~\cite{ArnoldEtAl2005,BalaEtAl2000}. Amdahl's law further quantifies this limitation~\cite{Amdahl1967}: eliminating the execution cost of a region accounting for a runtime fraction \(p\) yields a speedup of at most \(1/(1-p)\). \ourtool{} therefore localizes project-owned \emph{hot functions} for subsequent optimization.

To identify these functions, \ourtool{} profiles each performance workload using \texttt{perf}, collecting cycle samples together with their call stacks, and attributes each sample to a project-owned function. Attribution by machine-code symbol is unreliable for translated code: small project functions are often inlined into their callers, including the harness, and some project functions delegate their work to C library routines. In both cases, the samples would be charged to code other than the function that should be rewritten. \ourtool{} therefore expands inlined frames using debug information and attributes each sample to the deepest project-owned frame on its call stack. For samples in project-owned code, attribution to the deepest project-owned frame prevents callers from appearing hot because of expensive project callees. Samples in non-project code, such as C library routines, are instead attributed to the invoking project function, which is the actionable rewrite target.
Let \(s(f,w)\) denote the fraction of cycle samples attributed to project-owned function \(f\) under workload \(w\), referred to as its \emph{attributed cycle share}. Because hotspots may vary across workloads, \ourtool{} defines project-level hotness as \(h(f)=\max_{w\in\mathcal{W}_{\mathrm{perf}}}s(f,w)\), where \(\mathcal{W}_{\mathrm{perf}}\) is the set of performance workloads. Taking the maximum preserves workload-specific hotspots that would be diluted by averaging across workloads. \ourtool{} ranks functions by \(h(f)\), discards those below the hotness threshold, and retains up to a fixed budget of the highest-ranked functions, yielding \(\mathcal{H}=\{f_1,f_2,\ldots,f_n\}\).

\subsubsection{Issue Pattern Matching}
\label{sec:detection}

Hotspot localization identifies which functions should be prioritized for optimization, but not the concrete performance issues within them. We encode each performance issue identified in RQ3~(\S\ref{sec:rq3}) as an
\emph{optimization rule} pairing its \emph{Observed Pattern} with the corresponding \emph{Optimization Direction}. The former defines the matching criteria for identifying issue occurrences, while the latter guides subsequent rewriting. \ourtool{} realizes the matching criteria through three analyzers over optimized IR, compiler remarks, and Rust source. Although some patterns are derived using paired C--Rust evidence, matching relies only on Rust-side artifacts.

\emph{Optimized IR Analyzer.}
\ourtool{} compiles the project to optimized LLVM IR with debug information and searches for rule-specific IR signatures. Surviving \texttt{panic\_bounds\_check} calls are candidate matches for \emph{Residual Bounds Checks}, while \texttt{llvm.fptosi.sat} and \texttt{llvm.fptoui.sat} intrinsics are candidate matches for \emph{Redundant Conversion Saturation}.

\emph{Optimization Remarks Analyzer.}
\ourtool{} compiles the project with optimization remarks enabled and matches remarks by their reporting pass and reason against the corresponding \emph{Observed Pattern}. Only remarks whose reported barriers conform to the corresponding \emph{Observed Pattern} are retained. For example, vectorization remarks reporting insufficient profitability are excluded, while inlining matches additionally require the call to remain in optimized IR. GVN and LICM remarks attributing failed load elimination or hoisting to potentially clobbering stores are candidate matches for \emph{Obscured Memory Access Properties}.

\emph{Rust Source Analyzer.}
\ourtool{} parses the project into syntax trees and builds a project-wide symbol index resolving types, globals, functions, and caller--callee relationships. These representations enable structural matching of the source-level \emph{Observed Patterns}. For example, a buffer that is repeatedly allocated and released or grown by fixed increments through \texttt{realloc} is a candidate match for \emph{Inefficient Buffer Management}.

Applying these analyzers yields candidate occurrences of the identified performance issues, each associated with a matched rule and a source location. Because the three analyzers report locations in different forms, \ourtool{} normalizes them before further processing: IR matches are mapped through debug metadata and inline-call chains, optimization remarks provide compiler-reported locations, and source matches directly identify syntax-tree ranges. \ourtool{} then converts these locations to project-relative source ranges, associates each match with its enclosing source function, and retains only matches within the hot-function set~\(\mathcal{H}\).
To remove duplicate matches and form coherent rewrite regions, \ourtool{} consolidates the retained matches in two stages. First, matches that associate the same rule with the same source construct, including those from different analyzers, are merged into a single \emph{pattern instance}. Second, co-located instances are combined into a \emph{pattern group} when their source regions overlap, nest, or occur in nearby statements of the same block, subject to a bound on the enclosing region size; otherwise, an instance forms a singleton group. Each group is anchored at the smallest enclosing syntactic region covering its instances, such as an expression, a statement sequence, or a loop.

For each hot function \(f \in \mathcal{H}\), \ourtool{} represents the resulting groups as
\(
M_f = \{\langle \ell_i, R_i \rangle\}_{i=1}^{m_f},
\)
where \(m_f\) is the number of candidate optimization regions in \(f\),
\(\ell_i\) is the source region of the \(i\)-th group, and \(R_i\) is its set
of matched optimization rules. Each pair therefore specifies where to rewrite
and which optimization directions to consider jointly. For example, a residual
bounds-check match at an indexed expression and a vectorization-failure match
at its enclosing loop may be grouped at the loop level. The resulting group
contains both \emph{Residual Bounds Checks} and \emph{Missed Loop
Vectorization}, allowing their optimization directions to be coordinated in a
single rewrite. These groups form the units for subsequent optimization.

\subsection{LLM-Based Rewrite Generation}
A pattern group localizes a source region and associates it with one or more matched optimization rules. A match, however, only indicates that the corresponding \emph{Optimization Direction} may be relevant; applying that direction requires rewrite-specific semantic preconditions that the \emph{Observed Pattern} alone does not establish. To make these requirements explicit, \ourtool{} augments each rule with \emph{rewrite preconditions}. These preconditions, together with the \emph{Observed Pattern} and \emph{Optimization Direction}, form an \emph{optimization card}.
The LLM agent then evaluates these preconditions at the matched region before synthesizing a candidate rewrite.

\begin{figure*}[t]
    \centering
    \includegraphics[width=\textwidth]{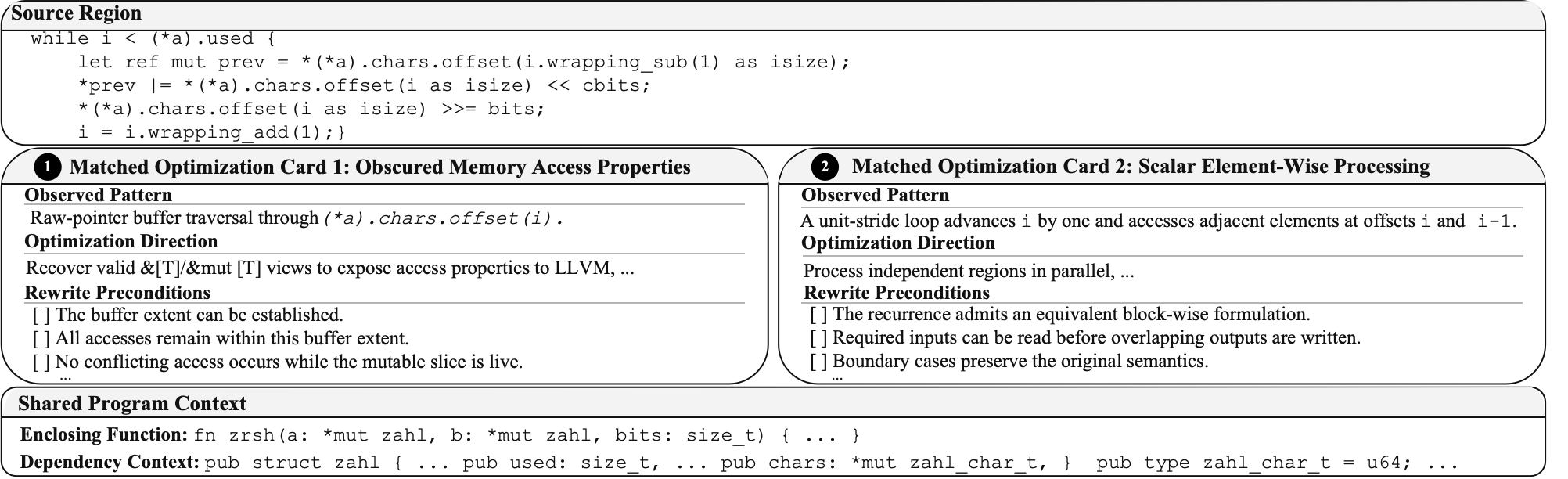}
    \caption{Region-specific optimization prompt for the \texttt{zrsh} loop}
    \label{fig:optimization-prompt}
\end{figure*}

For each concrete rewrite form associated with an \emph{Optimization Direction}, we identify the semantic obligations required to preserve the behavior of the original code. These obligations follow from established legality conditions for the corresponding program transformation where applicable~\cite{DBLP:books/mk/AllenK2001, DBLP:books/mk/Muchnick1997}; otherwise, they are derived from the behavioral-equivalence requirements of the rewrite. When a rewrite introduces a Rust construct or API, we additionally incorporate its documented validity requirements~\cite{rust-std}. For example, Figure~\ref{fig:optimization-prompt} illustrates mutable-slice recovery for \emph{Obscured Memory Access Properties}: the corresponding card requires establishing the buffer extent, ensuring that accesses remain within that extent, and excluding conflicting accesses while the mutable slice is live. We express these obligations as concrete checks over the target code, forming the card's \emph{rewrite-precondition checklist}.

To enable the LLM to assess the rewrite-precondition checklist, \ourtool{} provides the program context needed to establish the required conditions. For each pattern group \(\langle\ell_i,\mathcal{R}_i\rangle\in M_f\), \ourtool{} instantiates the associated optimization cards as \emph{matched optimization cards} by grounding each \emph{Observed Pattern} in the corresponding source construct. The enclosing function provides the local control- and data-flow context. When a precondition depends on information beyond the function, \ourtool{} retrieves relevant type, global, and function definitions, together with project-local call sites, through the project-wide symbol index as \emph{dependency context}. The source region \(\ell_i\), matched cards, and collected context together form a region-specific optimization prompt. Figure~\ref{fig:optimization-prompt} illustrates this construction for the \texttt{zrsh} loop, where the definitions of \texttt{zahl} and \texttt{zahl\_char\_t} provide dependency context for \texttt{(*a).chars}.
Based on the region-specific prompt, the agent first evaluates the rewrite preconditions of each matched card against the supplied program context. Cards with unmet or unresolved preconditions are excluded; if no card remains, \ourtool{} skips the region. Otherwise, the agent jointly considers the \emph{Optimization Directions} of the applicable cards and synthesizes a single \emph{candidate rewrite} for \(\ell_i\). When multiple directions affect overlapping code, they are coordinated within the same rewrite rather than applied independently.

\ourtool{} applies this procedure sequentially to all pattern groups, processing
hot functions in descending order of \(h(f)\). Each group is optimized against
the current accepted project version, so previously accepted rewrites are
reflected in subsequent optimization. After validation
(\S\ref{sec:validation}), an accepted rewrite becomes the new project version;
otherwise, the current version is retained. Because accepted rewrites may shift
source locations or alter dependencies, \ourtool{} re-resolves the remaining
groups against the updated project before continuing.

\subsection{Execution-based Rewrite Validation}
\label{sec:validation}
Applying a candidate rewrite may cause either (i)~behavioral deviation or (ii)~runtime regression despite preserving behavior. \ourtool{} therefore validates every candidate for both functionality and performance before acceptance. The refined Rust project serves as the prepared baseline and initializes the current accepted version. A candidate is applied to this version and becomes the new accepted version only if it passes both validations; otherwise, the previous version is restored.

\parabf{Functional Validation.} Before rewriting begins, \ourtool{} replays the functional workloads from \S\ref{sec:workload} on the prepared baseline to verify their recorded reference digests. For each candidate, \ourtool{} rebuilds the project and re-executes the same operation--input pairs with the same harnesses. A candidate passes functional validation only if the build succeeds, all executions complete normally, and every output digest matches its reference. Otherwise, \ourtool{} invokes a feedback-driven repair loop following prior LLM-based program repair~\cite{DBLP:journals/pacmse/Yuan0DW00L24, ptrTrans}. The available diagnostics and source locations, together with the attempted rewrite and relevant program context, are returned to the agent for revision. The revised candidate is rebuilt and revalidated until it passes or the repair budget is exhausted, after which the candidate is discarded.

\parabf{Performance Validation.} An \emph{Optimization Direction} does not guarantee an end-to-end speedup, as semantics-preserving rewrites may alter compiler optimizations and machine-code layout~~\cite{DBLP:journals/pacmpl/GongCS0SWMPVNT18, DBLP:conf/asplos/CurtsingerB13}. \ourtool{} therefore evaluates each functionally valid candidate using two complementary performance gates. Measurements use paired runs of the candidate and the corresponding reference version on each performance workload, with execution order alternated across pairs. Confidence intervals are computed for per-operation and aggregate runtime changes, and inconclusive cases receive additional measurements within a predefined budget. The \emph{incremental gate} compares the candidate against the current accepted version on workloads associated with the rewritten hot function, ensuring that the new rewrite does not introduce a statistically supported regression beyond the predefined tolerance. Its aggregate decision weights operations by the hot function's execution contribution. The \emph{cumulative gate} compares the candidate against the prepared baseline across all measurable performance workloads, ensuring that successive accepted rewrites continue to provide a project-wide net gain. It uses an equal-weight aggregate and additionally guards against severe operation-level regressions that could be hidden by the aggregate result.
Candidates with unresolved decisions or measurement failures at budget exhaustion are rejected.

\section{Evaluation}

\begin{table*}[t]
    \centering
\caption{Per-operation runtime overhead over C on the Evaluation Dataset.
Positive values indicate slowdown and negative values speedup; n/e denotes
non-equivalent output.}

    \label{tab:evaluation-performance}

    \scriptsize
    \setlength{\tabcolsep}{1.8pt}
    \renewcommand{\arraystretch}{1.03}
\begin{adjustbox}{width=\textwidth}
    \begin{tabular}{
      @{}
      l >{\raggedright\arraybackslash}p{1.03in} r r r r r
      @{\hspace{0.8em}}
      l >{\raggedright\arraybackslash}p{1.03in} r r r r r
      @{}
    }
      \toprule

      \textbf{Project}
      & \textbf{Operation}
      & \textbf{\raw{}}
      & \textbf{\codex{}}
      & \textbf{\refined{}}
      & \textbf{\ablation{}}
      & \textbf{\ourtool{}}
      &
      \textbf{Project}
      & \textbf{Operation}
      & \textbf{\raw{}}
      & \textbf{\codex{}}
      & \textbf{\refined{}}
      & \textbf{\ablation{}}
      & \textbf{\ourtool{}}
      \\

      \midrule

      \textsc{libopenaptx}
      & aptX encoding
      & $+3.7\%$ & n/e & $+3.7\%$ & $-5.4\%$ & $\mathbf{-9.0\%}$
      &
      \textsc{zopfli}
      & DEFLATE compression
      & $-25.9\%$ & $-25.9\%$ & $-25.2\%$ & $-24.9\%$ & $\mathbf{-31.4\%}$
      \\

      & aptX decoding
      & $+15.5\%$ & n/e & $+15.8\%$ & $+5.1\%$ & $\mathbf{-1.9\%}$
      &
      \textsc{lodepng}
      & PNG decoding
      & $+10.6\%$ & $+5.4\%$ & $+11.3\%$ & $\mathbf{+4.9\%}$ & $+5.0\%$
      \\

      & Stream recovery
      & $+9.6\%$ & n/e & $+9.7\%$ & $+0.1\%$ & $\mathbf{-5.6\%}$
      &

      & PNG encoding
      & $+0.6\%$ & $-10.8\%$ & $+0.8\%$ & $+0.2\%$ & $\mathbf{-15.0\%}$
      \\

      \textsc{http-parser}
      & Request parsing
      & $+141.8\%$ & $+124.1\%$ & $+168.2\%$ & $+167.5\%$ & $\mathbf{+10.1\%}$
      &

      & Color conversion
      & $-0.1\%$ & $+9.7\%$ & $+31.8\%$ & $+29.2\%$ & $\mathbf{-45.5\%}$
      \\

      & Response parsing
      & $+294.8\%$ & $+235.3\%$ & $+350.0\%$ & $+295.0\%$ & $\mathbf{+12.3\%}$
      &

      & zlib processing
      & $+1.4\%$ & $-7.9\%$ & $+1.1\%$ & $+0.3\%$ & $\mathbf{-25.0\%}$
      \\

      & Incremental parsing
      & $+188.9\%$ & $+188.1\%$ & $+211.5\%$ & $+211.0\%$ & $\mathbf{+11.7\%}$
      &

      & Chunk processing
      & $+3.4\%$ & $\mathbf{-0.6\%}$ & $-0.3\%$ & $-0.4\%$ & $-0.3\%$
      \\

      & URL parsing
      & $-0.3\%$ & $+1.2\%$ & $+3.2\%$ & $+3.3\%$ & $\mathbf{-28.5\%}$
      &
      \textsc{libqrencode}
      & QR encoding
      & $+14.8\%$ & $+16.5\%$ & $+18.0\%$ & $+13.0\%$ & $\mathbf{+11.2\%}$
      \\

      \textsc{fzy}
      & Fuzzy scoring
      & $+32.2\%$ & $+14.8\%$ & $+32.4\%$ & $+31.4\%$ & $\mathbf{+13.5\%}$
      &

      & Micro-QR encoding
      & $+1.1\%$ & n/e & $+1.6\%$ & $+1.6\%$ & $\mathbf{+0.6\%}$
      \\

      & Choice-list search
      & $+51.0\%$ & $+44.9\%$ & $+52.0\%$ & $+50.7\%$ & $\mathbf{+36.1\%}$
      &

      & Structured append
      & $+16.2\%$ & $+19.0\%$ & $+19.4\%$ & $+14.5\%$ & $\mathbf{+13.0\%}$
      \\

      & Candidate-list parsing
      & $+50.9\%$ & $+44.0\%$ & $+51.6\%$ & $+50.0\%$ & $\mathbf{+38.3\%}$
      &
      \textsc{optipng}
      & PNG optimization
      & $+1.7\%$ & $+1.7\%$ & $+1.7\%$ & $+1.7\%$ & $\mathbf{+1.5\%}$
      \\

      \textsc{lil}
      & Script execution
      & $-1.5\%$ & $-5.6\%$ & $-1.7\%$ & $-2.0\%$ & $\mathbf{-9.4\%}$
      &

      & Format conversion
      & $+1.4\%$ & $\mathbf{+0.6\%}$ & $+1.0\%$ & $+0.9\%$ & $\mathbf{+0.6\%}$
      \\

      & Embedding API
      & $+11.1\%$ & $\mathbf{-6.3\%}$ & $+12.6\%$ & $+11.7\%$ & $-1.7\%$
      &
      \textsc{brotli}
      & Compression
      & $+3.7\%$ & $\mathbf{+1.6\%}$ & $+3.7\%$ & $+3.6\%$ & $+4.0\%$
      \\

      \textsc{miniz}
      & Compression
      & $+2.3\%$ & $+2.3\%$ & $+2.4\%$ & $+2.7\%$ & $\mathbf{+1.5\%}$
      &

      & Decompression
      & $+13.4\%$ & $+13.4\%$ & $+14.7\%$ & $+14.9\%$ & $\mathbf{+9.6\%}$
      \\

      & Decompression
      & $+25.6\%$ & $\mathbf{+7.3\%}$ & $+25.4\%$ & $+25.2\%$ & $+12.5\%$
      &

      & Streaming
      & $+4.1\%$ & $+4.0\%$ & $+2.8\%$ & $+2.7\%$ & $\mathbf{+2.2\%}$
      \\

      & Streaming
      & $+12.7\%$ & $\mathbf{+6.4\%}$ & $+12.6\%$ & $+12.7\%$ & $+6.8\%$
      &

      & Dictionary modes
      & $+6.5\%$ & $+6.6\%$ & $+6.6\%$ & $+6.2\%$ & $\mathbf{+5.7\%}$
      \\

      & Checksums
      & $+16.7\%$ & $+16.7\%$ & $+16.7\%$ & $+16.7\%$ & $\mathbf{-66.0\%}$
      &
      \textsc{libxml2}
      & Parsing
      & $+2.1\%$ & $+2.1\%$ & $+1.9\%$ & $\mathbf{+1.2\%}$ & $+1.4\%$
      \\

      & ZIP processing
      & $+4.9\%$ & $+2.3\%$ & $+4.9\%$ & $+5.1\%$ & $\mathbf{-28.1\%}$
      &

      & Validation
      & $+1.0\%$ & $+0.8\%$ & $-0.1\%$ & $-0.1\%$ & $\mathbf{-0.6\%}$
      \\

      \textsc{lz4}
      & Compression
      & $+3.2\%$ & $+3.3\%$ & $\mathbf{+1.5\%}$ & $\mathbf{+1.5\%}$ & $+1.6\%$
      &

      & XPath Evaluation
      & $+0.5\%$ & $+0.6\%$ & $-1.1\%$ & $\mathbf{-1.5\%}$ & $-0.9\%$
      \\

      & Decompression
      & $+9.8\%$ & $+17.5\%$ & $+14.2\%$ & $+14.1\%$ & $\mathbf{+6.9\%}$
      &

      & Serialization
      & $-1.1\%$ & $-1.4\%$ & $\mathbf{-2.1\%}$ & $-1.6\%$ & $-1.7\%$
      \\

      & Streaming
      & $+5.5\%$ & $+8.8\%$ & $+2.4\%$ & $+2.4\%$ & $\mathbf{-1.0\%}$
      &

      & Dictionary/URI/string utilities
      & $+13.1\%$ & $+11.4\%$ & $\mathbf{+6.3\%}$ & $+7.1\%$ & $+7.6\%$
      \\

      & Dictionary modes
      & $+3.1\%$ & $+5.6\%$ & $+1.4\%$ & $+1.4\%$ & $\mathbf{-1.7\%}$
      &

      &
      & & & & &
      \\

      \bottomrule
    \end{tabular}

    \end{adjustbox}
  \end{table*}

\label{sec:evaluation}

We evaluate \ourtool{} on the Evaluation Dataset to answer two research
questions:
\textbf{RQ4 (Optimization Effectiveness)}: How effectively does \ourtool{}
reduce the performance gap between translated Rust and C, and how does it
compare with a general-purpose coding agent?
\textbf{RQ5 (Ablation Study)}: What contributes to the performance improvement of \ourtool{}, and what is the contribution of rule guidance?

\subsection{Evaluation Setup}
\label{sec:eval-setup}

\parabf{Implementation.}
\ourtool{} combines AST-based Rust source analysis with a project-wide symbol index, optimized LLVM IR and optimization-remark analysis, and \texttt{perf}-based cycle profiling. All LLM-based components use GPT-5.4~\cite{chatgpt}. Workload synthesis targets 70\% public-function coverage, with at most three coverage-refinement rounds and five harness-repair attempts. Hotspot localization retains project-owned functions with project-level hotness \(h(f)\) of at least 3\%. Internal performance validation uses two-sided 95\% confidence intervals with adaptive sampling at 10, 20, and 40 measurement pairs. The per-operation and aggregate regression tolerances are 1.0\% and
0.3\%, respectively, and accepted rewrites must provide at least 0.1\% project-wide net gain.

\parabf{Baselines.}
We use the C implementation as the runtime reference and \raw{} as the
unoptimized Rust baseline. As an external baseline, we use \codex{}~\cite{OpenAI:CodexCLI}, a general-purpose coding agent, and instruct it to optimize each \raw{} project for runtime performance. For RQ5, we additionally consider \refined{}, the output of Pre-optimization Normalization~(\S\ref{sec:translation-refinement}), and \ablation{}, which omits optimization cards from rewrite-generation prompts while retaining the rest of the pipeline.

\parabf{Evaluation Settings.}
We use the Evaluation Dataset workloads, build configuration, execution environment, measurement protocol, and runtime-overhead metric defined in \S\ref{sec:benchmarks} and \S\ref{sec:setup}. To control code-placement effects in per-operation comparisons, we additionally compile all variants, including C, with 64-byte function alignment. We run \codex{} using CLI v0.153.4 with the same GPT-5.4 model as \ourtool{} and reasoning effort set to \texttt{high}. For each project, \codex{} receives the complete \raw{} project and is instructed to optimize runtime performance while preserving program behavior, with a 30-minute budget. We evaluate all variants on the same workloads and report timings only for operation--input pairs whose outputs are identical to those of C.

\subsection{RQ4: Optimization Effectiveness}
\label{sec:rq4}

Table~\ref{tab:evaluation-performance} reports the per-operation runtime overhead relative to C for \raw{}, \codex{}, and \ourtool{} across 41 operations from 12 projects. Overall, \ourtool{} reduces measured runtime relative to \raw{} on 40 of the 41 operations. The improvement is particularly large for \textsc{http-parser}: across its three message-parsing operations, overhead decreases from as high as $294.8\%$ for \raw{} to at most $12.3\%$ for \ourtool{}. After optimization, 18 operations run faster than C, including 10 by more than 3\%. The largest gain relative to C occurs on \textsc{miniz} checksum computation, where overhead changes from $+16.7\%$ for \raw{} to $-66.0\%$ for \ourtool{}.

Compared with \codex{}, \ourtool{} preserves output equivalence on more operations and achieves lower runtime on most comparable cases. \codex{} produces non-equivalent outputs on four operations, whereas \ourtool{} remains output-equivalent to C on all 41. Among the 37 operations for which both produce equivalent outputs, \ourtool{} achieves lower measured runtime on 31, compared with five for \codex{}, with one tie; its largest speedup over \codex{} reaches $3.4\times$ on \textsc{miniz} checksum computation.

\finding{\ourtool{} effectively recovers translation-induced performance loss,
achieving competitive or better performance than C while preserving behavior
and outperforming a general-purpose coding agent overall.}

\subsection{RQ5: Ablation Study}
\label{sec:rq5}

Table~\ref{tab:evaluation-performance} compares \refined{}, \ablation{}, and \ourtool{} to assess the contribution of rule guidance. Translation Refinement and unguided optimization can both improve runtime on individual operations, but their gains are limited compared with \ourtool{}.  Overall, \ourtool{} achieves lower measured runtime than \ablation{} on 34 of the 41 operations. For example, on \textsc{lz4} streaming and dictionary modes, overhead decreases from $+5.5\%$ and $+3.1\%$ for \raw{} to $+2.4\%$ and $+1.4\%$ after refinement. These values remain unchanged under \ablation{}, whereas \ourtool{} further reduces them to $-1.0\%$ and $-1.7\%$, respectively. A larger gap appears for \textsc{lodepng} zlib processing, where overhead is $+1.1\%$ after refinement and $+0.3\%$ under \ablation{}, but drops to $-25.0\%$ under \ourtool{}. These results indicate that the derived rules provide effective guidance for recovering performance beyond what refinement and unguided optimization can achieve.

To understand how rule guidance contributes to these gains, we further examine the rewrites retained after validation. \ourtool{} yields 62 accepted rewrites, compared with 28 for \ablation{}. The two variants also differ in the types of transformations they produce. Among the 28 rewrites accepted under \ablation{}, 16 are local cleanups; none recovers additional slice views or removes residual bounds checks, leaving the raw-pointer accesses remaining after refinement unchanged. In contrast, 46 of the 62 rewrites accepted by \ourtool{} involve \emph{Memory Property Recovery}, making access relationships and invariants explicit to LLVM. These results indicate that rule guidance changes not only the number but also the nature of accepted rewrites, directing optimization toward the performance mechanisms identified in our empirical study.

\finding{The optimization rules distilled from our study effectively
guide the recovery of translation-induced performance loss, substantially
outperforming unguided optimization.}

\section{Discussion}
\label{sec:discussion}

\subsection{Sources of Performance Improvement}
\label{sec:disc-cost}
Among the 41 Evaluation-Dataset operations, \raw{} incurs more than 3\% overhead over C on 27. We apply the RQ2 decomposition to these regressions. Among them, 16 execute more instructions at lower CPI, nine increase both instruction count and CPI, and two increase CPI alone. Thus, instruction inflation appears in 25 of 27 regressions and remains the predominant overhead source.

Across the 27 regressions, \ourtool{} reduces the geometric mean of
\(R_{\mathrm{inst}}\) from 1.287 to 1.012, while \(R_{\mathrm{CPI}}\) remains near unity (0.987 to 1.003), indicating that performance recovery primarily comes from eliminating translation-induced extra instructions. Residual slowdowns, however, can still be CPI-dominated. The two largest, \textsc{fzy} choice-list search and candidate-list parsing, execute fewer instructions than C after optimization but remain slower because of higher CPI, with TMA attributing most of the gap to Core Bound and Memory Bound.

Despite this overall trend, individual rule-guided rewrites can also reduce per-instruction cost. In \textsc{miniz} checksum computation, a rewrite for \emph{Scalar Element-Wise Processing} replaces byte-wise processing with an equivalent block-wise formulation. The optimized code executes nearly the same number of instructions as C (\(R_{\mathrm{inst}}=0.986\)) but at roughly one third of its CPI (\(R_{\mathrm{CPI}}=0.344\)); TMA likewise shows Memory Bound decreasing from 47\% in C to 11\% under \ourtool{}. This case shows that rule-guided rewriting can improve efficiency not only by reducing extra work, but also by lowering per-instruction cost.

\subsection{Cross-Project Applicability of Optimization Rules}
\label{sec:disc-rules}
The optimization rules operationalized from the RQ3 findings recur broadly across the 12 disjoint Evaluation projects. At least one rule-specific pattern is detected in 94\% of the hot functions; every project exhibits patterns associated with at least three rules (median: four), and all nine rules contribute to accepted rewrites. Together, these results indicate that the optimization knowledge identified in the Study Dataset remains applicable beyond the projects used for its derivation.

The prevalence of matched patterns varies substantially across rules.
\emph{Obscured Memory Access Properties} is the most widespread, matching
89\% of the hot functions across all 12 projects and contributing to accepted rewrites in 11. \emph{Byte-Oriented Memory Operations} is also common, matching 35\% of the hot functions across 11 projects. Other source-level patterns occur less frequently, reflecting their narrower structural conditions. Compiler-level issues are likewise detected across projects: \emph{Residual Bounds Checks}, \emph{Missed Hot-Callee Inlining}, and \emph{Missed Loop Vectorization} match 18\%, 16\%, and 24\% of the hot functions, respectively. Together, these results show that cross-project applicability spans both source-level patterns and compiler-level optimization losses.

\subsection{Impact on Safety}
\label{sec:disc-unsafe}

Beyond runtime performance, we examine the safety-related effects of
\ourtool{} across the 12 projects, considering both the caller-visible
\texttt{unsafe} interface and raw-pointer use within rewritten functions.
In \raw{}, $99.8\%$ of functions are declared \texttt{unsafe fn}; \codex{}
leaves this proportion unchanged, whereas \ourtool{} reduces it to $60.5\%$,
increasing the proportion of functions callable from safe Rust from $0.2\%$
to $39.5\%$. This narrower interface is established during Translation
Refinement and preserved by subsequent optimization. Relative to \refined{},
the optimized functions further reduce pointer-arithmetic operations by
$18\%$ and raw-pointer dereferences by $10\%$ through transformations such
as slice-based traversal and iterator-based access. Overall, \ourtool{}
improves performance while narrowing the caller-visible unsafe interface and
further reducing reliance on raw-pointer memory operations.

\section{Threats to Validity}
\label{sec:threats}

One threat concerns the sensitivity of runtime measurements to system noise. We mitigate this threat by using five warm-up rounds and 30 paired measurements for each workload, alternating the C--Rust execution order, and compiling all compared variants with consistent function alignment. Another threat concerns the representativeness of performance evaluation within each project. Rather than using a single whole-project runtime or isolated API calls, we evaluate semantically complete end-to-end operations that exercise distinct implementation paths and use multiple operations when a project contains different performance-relevant functionalities. A third threat concerns potential bias in the optimization rules toward the projects from which they are derived. We mitigate this threat by deriving the rules exclusively from the eight Study-Dataset projects and reserving twelve disjoint projects for evaluation.

\section{Related Work}
\label{sec:related}
Automated code translation spans both function- and project-level migration
across multiple language pairs~\cite{DBLP:conf/icse/PanIKSWMSPSJ24,
DBLP:journals/pacmse/Yang0YK0LHMJ024, DBLP:journals/corr/abs-2410-24117,
guan2025repotransagentmultiagentllmframework,
DBLP:journals/corr/abs-2412-08035, DBLP:conf/nips/RoziereLCL20,
10938485, DBLP:journals/corr/abs-2404-18852}.
We focus on project-level C-to-Rust translation, where existing approaches
broadly follow two routes.
LLM-based translators aim to generate safer and more idiomatic Rust,
increasingly leveraging program analysis and project-wide context to improve
translation quality~\cite{DBLP:journals/corr/abs-2501-14257,PR2,
DBLP:journals/corr/abs-2503-17741,ptrTrans}.
Preserving source behavior, however, remains challenging at project scale.
Rule-based translators such as Corrode~\cite{Corrode} and
\crust{}~\cite{c2rust} instead apply deterministic transformations to C
constructs, yielding reproducible translations intended to preserve source
behavior. Such translations often retain C-oriented representations,
including raw pointers and extensive \texttt{unsafe}
code~\cite{DBLP:journals/pacmpl/EmreSDH21}.
To improve C-to-Rust translation, prior work has explored program-analysis-based
refinement~\cite{DBLP:journals/pacmpl/EmreSDH21,
DBLP:journals/pacmpl/EmreBPSDH23,crown}, LLM-guided
rewriting~\cite{DBLP:journals/corr/abs-2501-14257,PR2}, and project-wide
reasoning over dependencies and pointer
semantics~\cite{DBLP:journals/corr/abs-2503-17741,ptrTrans}.
These efforts primarily target behavioral preservation, safety, and
idiomaticity. In contrast, \ourtool{} starts from mechanically translated Rust
and focuses on recovering runtime performance lost during translation.

Recent work has begun to investigate performance regressions introduced by
C-to-Rust translation. Xia et al.\ compare randomly generated C programs with
their \crust{} translations and identify Relooper-induced code duplication as
one source of runtime overhead~\cite{xia2023empirical}, while RUSTY targets
this control-flow expansion through a specialized translation
scheme~\cite{DBLP:conf/qrs/HanHWZ22}.
However, they either rely on synthetic programs or address a specific
translation-induced mechanism, leaving performance divergence and recurring
code-level causes in real-world projects less well characterized.
Beyond translation-specific performance
studies, SemOpt derives reusable optimization strategies from historical
commits~\cite{ZhaoEtAl2026SemOpt}, and MoST extends such knowledge across
multiple sources, languages, and scenarios~\cite{ZhaoEtAl2026MoST}. While
these approaches demonstrate the value of rule-guided optimization,
their guidance is general-purpose rather than derived from performance
differences introduced by code translation. Our work connects these two
directions by analyzing paired C--Rust performance in real-world projects,
distilling recurring code-level performance patterns and optimization
directions into reusable rules, and applying them to guide post-translation
performance recovery.

\section{Conclusion}
\label{sec:conclusion}
Across 20 real-world projects, we studied performance degradation in automated
C-to-Rust translation and identified nine recurring performance issues and
optimization directions through paired C--Rust analysis. We then developed
\ourtool{}, a rule-guided framework for locating and optimizing these issues.
On twelve held-out projects, \ourtool{} preserves C outputs on all 41
operations and improves runtime over \crust{} on 40, demonstrating the value
of empirically derived knowledge for cross-project performance recovery.

\bibliographystyle{IEEEtran}

\bibliography{ref}

\end{document}